\documentclass[12pt]{article}
\usepackage{amsfonts}
\usepackage{amsmath}
\usepackage{amssymb}
\usepackage{amsthm}
\usepackage{geometry}
\usepackage{mathtools}
\usepackage{graphicx}
\usepackage{subfigure}
\usepackage{lineno,hyperref}
\modulolinenumbers[1]
\usepackage{mathrsfs}
\usepackage{url}
\usepackage{xcolor}
\usepackage{xspace}
\usepackage{stmaryrd}
\usepackage[ruled, linesnumbered, vlined]{algorithm2e}

\usepackage{multicol}

\newtheorem{theorem}{Theorem}

\newtheorem{example}[theorem]{Example}

\newtheorem{remark}[theorem]{Remark}

\renewcommand{\Pr}{\mathit{Pr}}

\newif\iftodo
\todotrue    

\newcommand{\ie}{i.e.}
\newcommand{\eg}{e.g.}
\newcommand{\nd}{\noindent}
\newcommand{\triple}[1]{(#1)}

\newcommand{\rhodf}{\mbox{$\rho$df}\xspace}

\newcommand{\subclass}{\ensuremath{\mathsf{sc}}}
\newcommand{\type}{\ensuremath{\mathsf{type}}}

\newcommand{\K}{{\cal K}}
\newcommand{\KB}{{\K}}

\newcommand{\wrt}{w.r.t.\xspace}

\newcommand{\I}{\ensuremath{\mathcal{I}}\xspace}      
\newcommand{\fuzzyg}[2]{\mbox{$\tuple{ #1, #2}$}}
\newcommand{\ins}{\,{\in}\,}
\newcommand{\notf}{\neg}
\newcommand{\orf}{\lor}
\newcommand{\andf}{\land}
\newcommand{\impf}{\Rightarrow}
\newcommand{\highi}[1]{{#1}^{\cal I} }
\newcommand{\unit}{[0,1]}
\newcommand{\ges}{\,{\geqslant}\,}

\newcommand{\eqs}{\,{=}\,}
\newcommand{\mids}{\,{\mid}\,}
\newcommand{\bsd}[2]{bsd(#1,#2)}
\newcommand{\bed}[2]{bed(#1,#2)}
\newcommand{\term}[1]{\ensuremath{\mbox{\small$#1$}}}
\newcommand{\assign}{\mbox{$\colon\!\!\!\!=$}}
\newcommand{\tuple}[1]{\langle #1 \rangle }
\renewcommand{\iff}{if and only if\xspace}

\newcommand{\andc}{\sqcap}
\newcommand{\all}{\forall}
\newcommand{\some}{\exists}

\newcommand{\impc}{\sqsubseteq}

\newcommand{\defc}{=}
\newcommand{\cass}[2]{\mbox{$#1$:$#2$}}
\newcommand{\rass}[3]{\mbox{$(#1,#2)$:$#3$}}

\newcommand{\lpx}[1]{{\mathcal #1}}
\newcommand{\lp}{\lpx{P}}

\newcommand{\cpl}[1]{\bar{#1}}
\newcommand{\liff}{\leftrightarrow}
\newcommand{\pfs}[1]{\tilde{2}^{#1}}

\newcommand{\reffig}[1]{Figure~\ref{#1}}

\newcommand{\luks}{\L ukasiewicz}
\newcommand{\godel}{G{\"{o}}del}

\newcommand{\limpf}{\rightarrow}
\newcommand{\mydef}{\overset{\mathtt{def}}{=}}

\newcommand{\cp}[1]{\mathsf{#1}}
\newcommand{\role}[1]{\mathsf{#1}}
\newcommand{\ind}[1]{\mathsf{#1}}

\newcommand{\myparagraph}[1]{\vspace{2ex} \noindent{\bf #1.}}

\newcommand{\aggr}{@}

\begin{document}

\title{An Introduction to Fuzzy \& Annotated Semantic Web Languages\thanks{This is an updated version of~\cite{Straccia17} and acts as accompanying material to my invited talk and slides at the 
2018 Artificial Intelligence International Conference (A2IC-18).}}
%
%
\author{Umberto Straccia\\
ISTI - CNR, Pisa, ITALY\\
{umberto.straccia@isti.cnr.it}\\ 
}
\date{}

\maketitle              

\begin{abstract}
We present the state of the art in representing and
reasoning with fuzzy knowledge in Semantic Web Languages such as triple
languages RDF/RDFS, conceptual languages of the OWL 2 family and rule
languages. We further show  how one may generalise them to so-called
annotation domains, that cover also e.g. temporal and provenance extensions.

\end{abstract}
\section{Introduction}

Managing uncertainty and fuzziness is growing in importance in Semantic Web research as recognised by a large number  of research efforts in this direction~\cite{Straccia08a,Straccia13,Straccia15a,Straccia17a,Straccia17}. \emph{Semantic Web Languages} (SWL) are the languages used to provide a formal description of concepts, terms, and relationships within a given domain, among which the \emph{OWL 2 family} of languages~\cite{OWL2}, \emph{triple languages}  RDF \& RDFS~\cite{RDFS} and  \emph{rule languages} (such as RuleML~\cite{ruleml}, Datalog$^{\pm}$~\cite{Cali09b} and RIF~\cite{RIF}) are major players.

While their syntactic specification is based on XML~\cite{XML}, their semantics is based on logical formalisms: briefly,
\begin{itemize}
\item RDFS is a logic having intensional semantics and the  logical counterpart is $\rhodf$~\cite{Munoz07};
\item OWL 2 is a family of languages that relate to \emph{Description Logics} (DLs)~\cite{Baader03a};
\item rule languages relate roughly to the \emph{Logic Programming} (LP) paradigm~\cite{Lloyd87};
\item both OWL 2 and rule languages have an extensional semantics.
\end{itemize}

 \myparagraph{Uncertainty versus Fuzziness} \label{serain}
 One of the major difficulties, for those unfamiliar on the topic, is to understand the conceptual differences between uncertainty and fuzziness.  Specifically, we recall that there has been a long-lasting misunderstanding in the literature of artificial intelligence and uncertainty modelling, regarding the role of probability/possibility theory and vague/fuzzy theory.  A clarifying paper is~\cite{Dubois01}. We recall here the salient concepts.

\paragraph{Uncertainty.} Under \emph{uncertainty theory} fall all those approaches in which statements rather than being either true or false, are true or false to some 
\emph{probability} or \emph{possibility}  (for example, ``it will rain tomorrow''). That is, a statement is true or false in any world/interpretation, but we are ``uncertain'' about which world to consider as the right one, and thus we speak about \eg~a probability distribution or a possibility distribution over the worlds. For example, we cannot exactly establish whether it will rain tomorrow or not, due to our \emph{incomplete} knowledge about our world, but  we can estimate to which degree this is probable, possible, or necessary. 

To be somewhat more formal, consider a propositional statement (formula) $\phi$ (``tomorrow it will rain") and a propositional interpretation (world) $\I$.  We may see $\I$ as a function mapping propositional formulae into $\{0,1\}$, \ie~$\I(\phi) \in \{0,1\}$. If $\I(\phi) = 1$, denoted also as $\I \models \phi$, then we say that the statement $\phi$ under $\I$ is true, false otherwise. Now, each interpretation $\I$ depicts some concrete world and, given $n$ propositional letters, there are $2^n$  possible interpretations. In uncertainty theory, we do not know which interpretation $\I$ is the actual one and we say that we are \emph{uncertain} about which world is the real one that will occur. 

To deal with such a situation, one may construct a \emph{probability distribution over the worlds}, that is a function $\Pr$ mapping interpretations in $[0,1]$, \ie~$\Pr(\I) \in [0,1]$, with $\sum_\I \Pr(\I) = 1$, where $\Pr(\I)$ indicates the probability that $\I$ is the actual world under which to interpret the propositional statement at hand.  Then, the {\em probability} of a statement~$\phi$ in $\Pr$, denoted $\Pr(\phi)$, is the sum of all $\Pr(\I)$ such that  $\I\models \phi$, \ie~
\[
\Pr(\phi) = \sum_{\I \models \phi} \Pr(\I) \ .
\]


\paragraph{Fuzziness.} On the other hand, under \emph{fuzzy theory} fall all those approaches in which statements  (for example, ``heavy rain'') are true to some  \emph{degree}, which is  taken from a truth space (usually $[0,1]$). 
That is, the convention prescribing that a proposition is either true or false is changed towards graded propositions.
For instance, the compatibility of ``heavy'' in the phrase ``heavy rain''  is graded and the degree depends on the amount of rain is falling.\footnote{More concretely, the intensity of precipitation is expressed in terms of a precipitation
rate $R$: volume  flux of precipitation through a horizontal surface, \ie~$m^3/ m^2 s = m s^{-1}$. It is usually expressed in $mm/h$.} Often we may find rough definitions about rain types, such as:~\footnote{\url{http://usatoday30.usatoday.com/weather/wds8.htm}}

\begin{description}
\item[Rain.] Falling drops of water larger than $0.5$ mm in diameter. In forecasts, ``rain" usually implies that the rain will fall steadily over a period of time;
\item[Light rain.] Rain falls at the rate of $2.6$ mm or less an hour;
\item[Moderate rain.] Rain falls at the rate of $2.7$ mm to $7.6$ mm an hour;
\item[Heavy rain.] Rain falls at the rate of $7.7$ mm an hour or more. 
\end{description}

\nd It is evident that such definitions are quite harsh and resemble a bivalent (two-valued) logic: \eg~a precipitation rate of $7.7mm/h$ is a heavy rain, while a precipitation rate of $7.6mm/h$ is just a moderate rain. This is clearly unsatisfactory, as quite naturally the more rain is falling, the more the sentence ``heavy rain''  is true and, vice-versa, the less rain is falling the less the sentence is true.
\begin{quote}
\emph{
 In other words, this means essentially, that the sentence ``heavy rain''  is no longer either true or false as in the definition above, but is intrinsically graded.
 }
\end{quote}

\begin{figure}
\begin{center}
\includegraphics[scale=0.6]{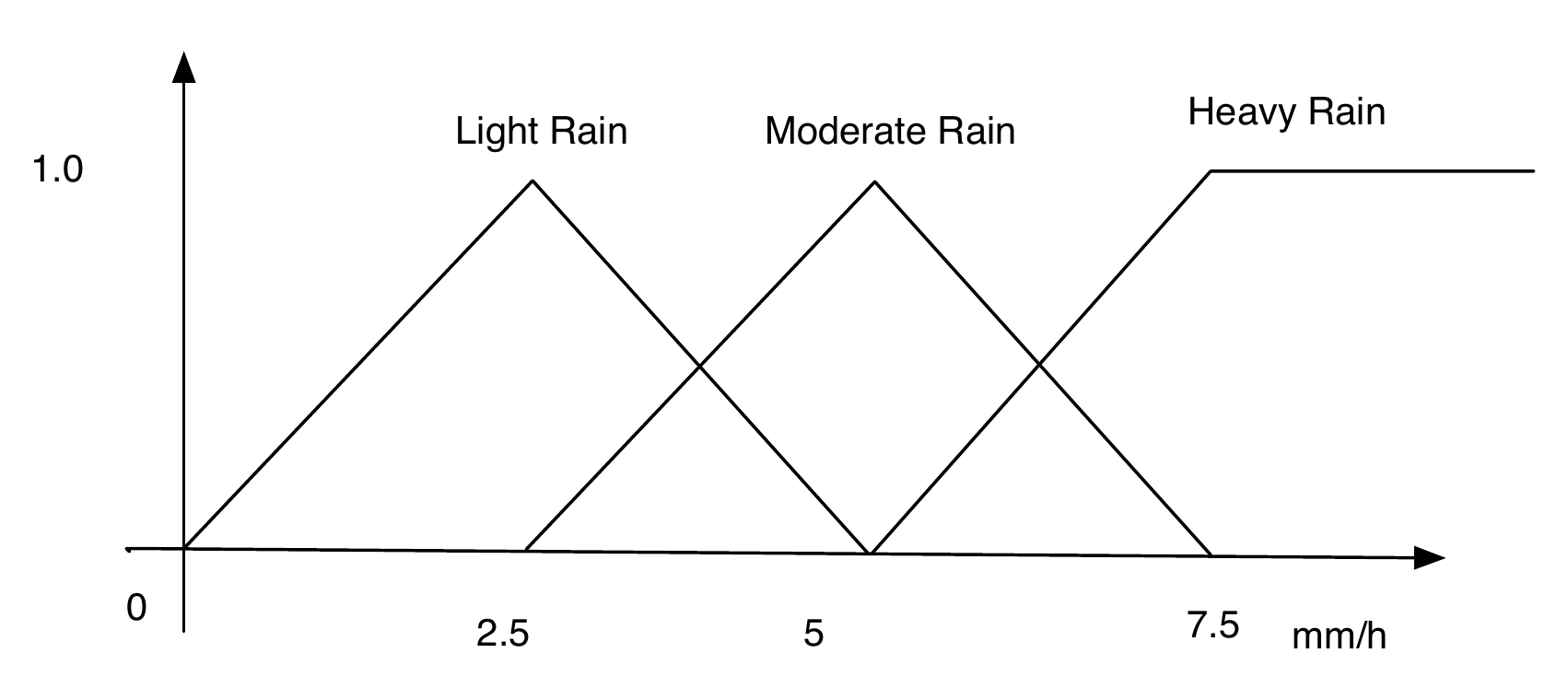}
\end{center}
\caption{Light, Moderate and Heavy Rain.} \label{rain}
\end{figure}

\nd A more fine grained way to define the various types of rains is illustrated in Figure~\ref{rain}.

\nd Light rain, moderate rain and heavy rain are called \emph{Fuzzy Sets} in the literature~\cite{Zadeh65} and are characterised by the fact that membership is a matter of degree. Of course, the definition of fuzzy sets is frequently context dependent and  subjective: \eg~the definition of heavy rain is quite different from heavy person and the latter may be defined differently among human beings.

From a logical point of view,  a propositional interpretation maps a statement $\phi$ to a truth degree  in $[0,1]$, \ie~$\I(\phi) \in [0,1]$. Essentially, we are unable to establish whether a statement is entirely true or false due to the involvement of \emph{vague/fuzzy} concepts, such as ``heavy''.

Note that all fuzzy statements are truth-functional, that is, the degree of truth of every statement can be calculated from the degrees of truth of its constituents, while uncertain statements cannot always be a function of the uncertainties of their constituents~\cite{Dubois94a}. For the sake of illustrative purpose, an example of truth functional interpretation of propositional statements is as follows:
\[
\begin{array}{lcl}
\I(\phi \andf \psi) & = & \min(\I(\phi), \I(\psi)) \\
\I(\phi \orf \psi) & = & \max(\I(\phi), \I(\psi))\\
\I(\notf \phi) & = & 1- \I(\phi) \ .
\end{array}
\]

\nd In such a setting one may be interested in the so-called notions of \emph{minimal (resp. maximal) degree of satisfaction} of a statement, \ie~$\min_\I \I(\phi)$ (resp.~$\max_\I \I(\phi)$).

\paragraph{Uncertain fuzzy sentences.}
Let us recap: in a probabilistic setting each statement is either true or false, but there is \eg~a probability distribution telling us how probable each interpretation is, \ie~$\I(\phi) \in \{0,1\}$ and $\Pr(\I) \in [0,1]$. In fuzzy theory instead, sentences are graded, 
\ie~we have $\I(\phi) \in [0,1]$.

A natural question is: can we have sentences combining the two orthogonal concepts? Yes, for instance, ``there will be heavy rain tomorrow" is an uncertain fuzzy sentence. Essentially, there is uncertainty about the world we will have tomorrow, and there is fuzziness about the various types of rain we may have tomorrow.

From a logical point of view, we may model uncertain fuzzy sentences in the following way:
\begin{itemize}
\item we have a probability distribution over the worlds, \ie~a function $\Pr$ mapping interpretations in $[0,1]$, \ie~$\Pr(\I) \in [0,1]$, with $\sum_\I \Pr(\I) = 1$;

\item sentences are graded. Specifically, each interpretation is truth functional and maps sentences into $[0,1]$, \ie~$\I(\phi) \in [0,1]$;

\item for a sentence $\phi$, we are interested in the so-called \emph{expected truth} of $\phi$, denoted $ET(\phi),$ namely
\[
ET(\phi) = \sum_\I Pr(\I)\cdot \I(\phi) \ .
\]
\end{itemize}

\nd Note that if $\I$ is bivalent (that is, $\I(\phi) \in\{0,1\}$) then $ET(\phi) = \Pr(\phi)$.

\paragraph{Overview.}
We present here some salient aspects in representing and reasoning with fuzzy knowledge in  Semantic Web Languages (SWLs) such as \emph{triple languages} ~\cite{RDFS} (see, \eg ~\cite{Straccia09f,Straccia10a}), \emph{conceptual languages}~\cite{OWL2} (see, \eg ~\cite{Lukasiewicz08a,Straccia01,Straccia06f}) and \emph{rule languages} (see, \eg~\cite{Damasio08,Ragone09,Straccia05b,Straccia05c,Straccia06c,Straccia06h,Straccia08a}).  We refer the reader to \cite{Straccia13} for an extensive presentation concerning fuzziness and  semantic web languages.
We then further show how one may generalise them to so-called annotation domains, that cover also e.g. temporal and provenance extensions (see, \eg~\cite{Lopes10,Lopes10a,Zimmermann12}).


\section{Basics: From Fuzzy Sets to Mathematical Fuzzy Logic and Annotation Domains } \label{sect:fuzzy}

\subsection{Fuzzy Sets Basics} \label{seccs}

The aim of this section is to introduce the basic concepts of fuzzy set  theory. To distinguish between fuzzy sets and classical (non fuzzy) sets, we refer to the latter as \emph{crisp sets}. For an in-depth  treatment we refer the reader to, \eg~\cite{Dubois80,Klir95}.

\paragraph{From Crisp Sets to Fuzzy Sets.}
To better highlight the conceptual shift from classical sets to fuzzy sets, we start with some basic definitions and well-known properties of classical sets. Let $X$ be a \emph{universal set} containing all possible elements of concern in each particular context. The \emph{power set}, denoted $2^{A}$, of a set $A \subset X$, is the set of subsets of $A$, \ie, $2^{A} = \{ B \mid B \subseteq A \}$. Often sets are defined by specifying a property satisfied by its members, in the form
$A = \{x \mid P(x) \}$, where $P(x)$ is a statement of the form ``$x$ has property $P$'' \emph{that is  either true or false} for any $x\in X$.  Examples of universe $X$ and subsets $A,B \in 2^{X}$ may be
\begin{eqnarray*}
X & = & \{x \mid \mbox{$x$ is a day} \} \\
A & = & \{x \mid \mbox{$x$ is a rainy day} \} \\
B & = & \{x \mid \mbox{$x$ is a day with  precipitation rate $R \geq 7.5mm/h$} \} \ .
\end{eqnarray*}
\nd In the above case we have $B \subseteq A \subseteq X$.

The \emph{membership function} of a set $A \subseteq X$, denoted $\chi_{A}$,   is a function mapping elements of $X$ into $\{0,1\}$, \ie~$\chi_{A}\colon X \to \{0,1\}$, where $\chi_{A}(x) = 1$ iff $x \in A$. Note that for any sets $A,B \in 2^{X}$, we have that 
\begin{equation} \label{ssincl}
A \subseteq B \mbox{ iff } \forall x \in X. \ \chi_{A}(x) \leq \chi_{B}(x) \ .
\end{equation}

\nd The \emph{complement} of a set $A$ is denoted $\cpl{A}$, \ie~$\cpl{A} = X \setminus A$.
Of course, $\forall x\in X. \ \chi_{\cpl{A}}(x) =  1 - \chi_{A}(x)$.
In a similar way, we may express set operations of intersection and union via the membership function as follows:
\begin{eqnarray}
\forall x\in X. \ \chi_{A \cap B}(x) & = & \min(\chi_{A}(x), \chi_{B}(x))  \label{sand} \\
\forall x\in X. \ \chi_{A \cup B}(x) & = & \max(\chi_{A}(x), \chi_{B}(x)) \ .\label{sor}
\end{eqnarray}

\nd The \emph{Cartesian product}, $A \times B$, of two sets $A, B \in 2^{X}$ is defined as 
$A \times B  = \{\tuple{a,b} \mid a \in A, b \in B \}$.  A relation  $R \subseteq X \times X$ is \emph{reflexive} if  for all $x\in X$
$ \chi_{R}(x,x)=1$, is \emph{symmetric} if  for all $x,y\in X$ $ \chi_{R}(x,y) =   \chi_{R}(y,x)$. The \emph{inverse} of $R$ is defined as 
function $ \chi_{R^{-1}}\colon X \times X \to \{0,1\}$ with membership function $ \chi_{R^{-1}}(y,x) =  \chi_{R}(x,y)$. 

As defined so far, the membership function of a crisp set $A$ assigns a value of either $1$ or $0$ to each individual of the universe set and, thus, discriminates between being a member or not  being a member of $A$.

A \emph{fuzzy set}~\cite{Zadeh65}   is characterised instead by a membership function $\chi_{A}\colon X \to \unit$, or denoted simply $A\colon X \to \unit$. With $\pfs{X}$ we denote the \emph{fuzzy power set} over $X$, \ie~the set of all fuzzy sets over $X$. For instance, by referring to Figure~\ref{rain}, the fuzzy set 
\begin{eqnarray*}
C & = & \{x \mid \mbox{$x$ is a day with  \emph{heavy} precipitation rate $R$} \} 
\end{eqnarray*}

\nd is defined via the membership function
\[
\chi_{C}(x) = 
\left \{
\begin{array}{ll}
1 & \mbox{if } R \geq 7.5 \\
(x -  5) / 2.5 & \mbox{if } R \in [5,7.5) \\
0 & \mbox{otherwise } \ .
\end{array} \right .
\] 

\begin{figure}
\begin{center}
\begin{tabular}{cc}
\includegraphics[scale=0.45]{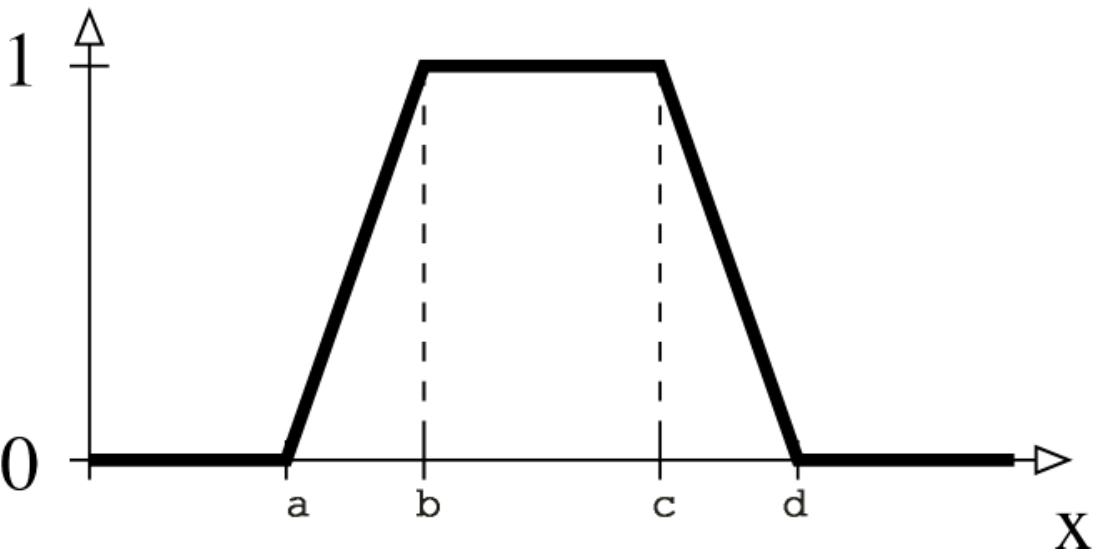}&
\includegraphics[scale=0.45]{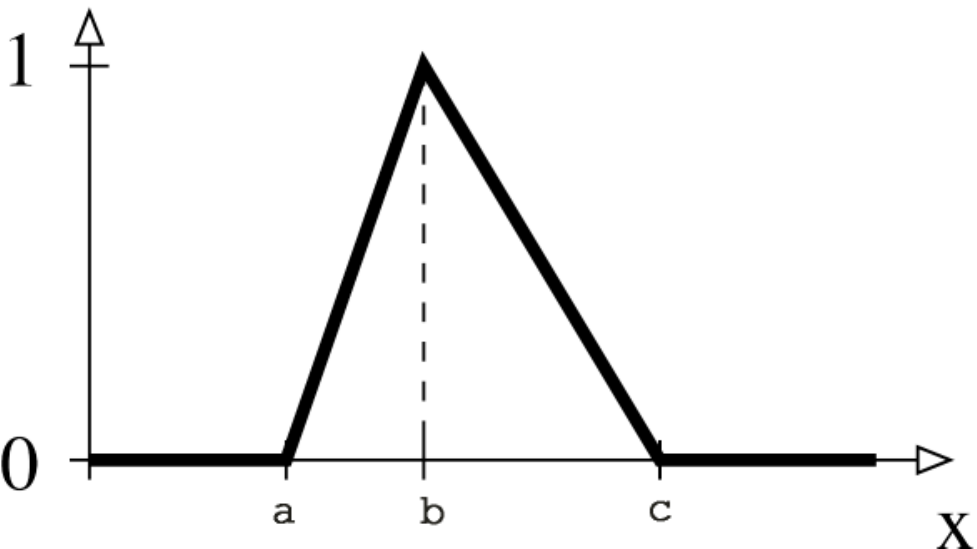} \\
(a) & (b) \\
\includegraphics[scale=0.45]{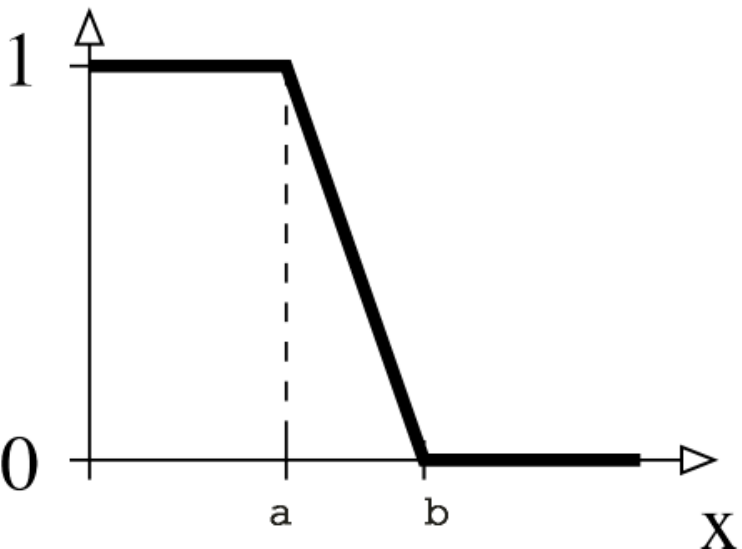} &
\includegraphics[scale=0.45]{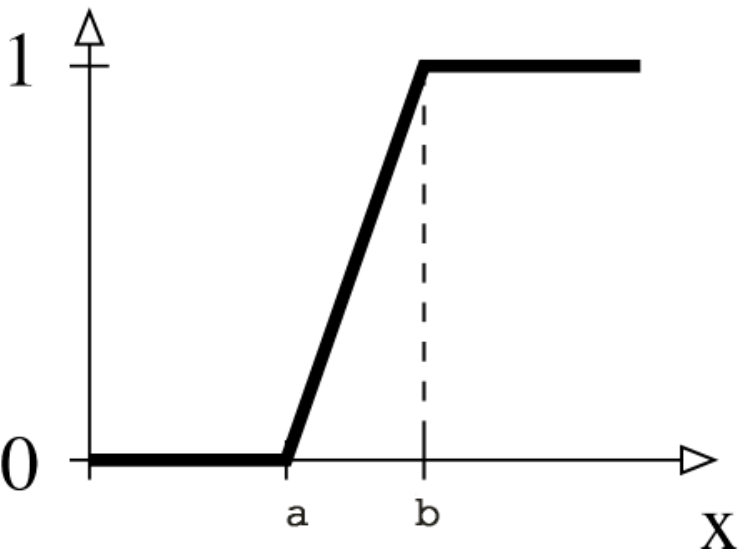} \\
(c) & (d)  \\
\end{tabular}
\caption{(a) Trapezoidal function $\mathit{trz}(a,b,c,d)$; (b) Triangular function $\mathit{tri}(a,b,c)$; (c)
$L$-function $\mathit{ls}(a,b)$; and (d) $R$-function $\mathit{rs}(a,b)$.}\label{muf}
\end{center}
\end{figure}

\nd As pointed out previously, the definition of the membership function may depend on the context and  may be subjective. Moreover, also the \emph{shape} of such functions may be quite  different. Luckily, the trapezoidal (Fig.~\ref{muf} (a)), the triangular (\reffig{muf} (b)), the $L$-function (left-shoulder function, \reffig{muf} (c)), and the $R$-function (right-shoulder function, \reffig{muf} (d)) are simple, but most frequently used to specify membership degrees. 

%
%

\begin{figure}
\begin{center}
\includegraphics[scale=0.5]{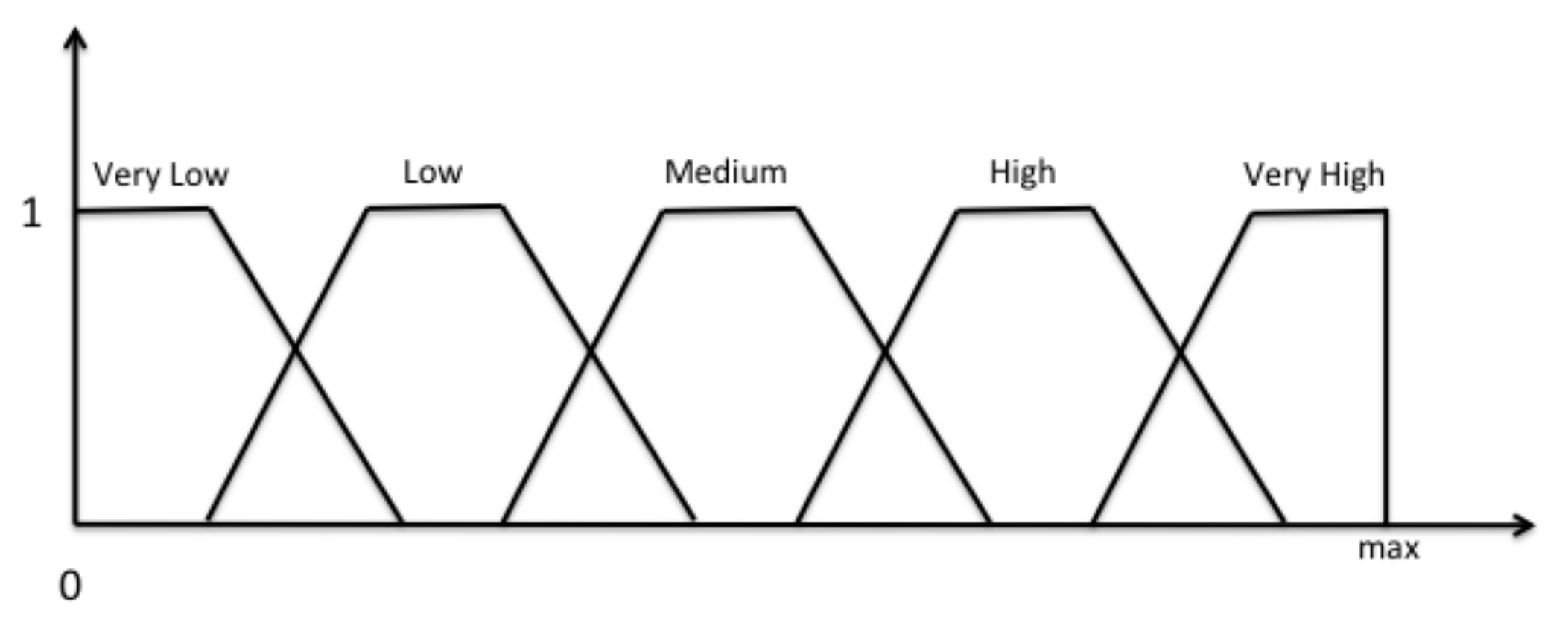}
\caption{Fuzzy sets construction using trapezoidal functions.}\label{partfuzzytrz}
\end{center}
\end{figure}

\begin{figure}
\begin{center}
\includegraphics[scale=0.5]{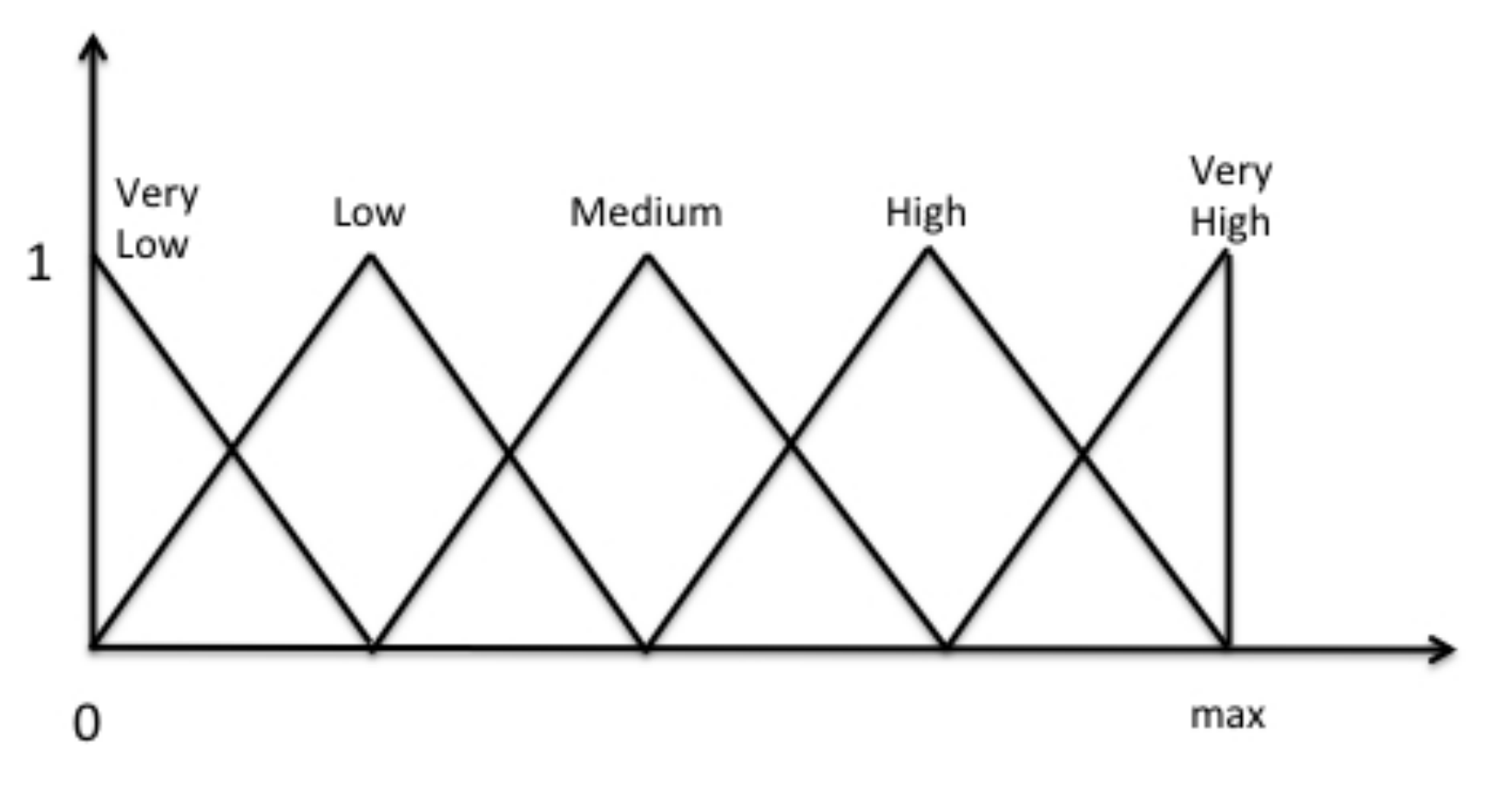}
\caption{Fuzzy sets construction using triangular functions.}\label{partfuzzytri}
\end{center}
\end{figure}

The usefulness of fuzzy sets depends critically on our capability to construct appropriate membership functions. The problem of constructing meaningful membership functions is a difficult one and we refer the interested reader to,~\eg~\cite[Chapter 10]{Klir95}. However, one easy and typically satisfactory method to define the membership functions (for a numerical domain) is to uniformly partition the range of, \eg~precipitation rates values (bounded by a minimum and maximum value), into 5 or 7 fuzzy sets using either trapezoidal functions (\eg~as illustrated in \reffig{partfuzzytrz}), or using triangular functions (as illustrated in \reffig{partfuzzytri}). The latter one is the more used one, as it has less parameters.

\begin{figure}
\begin{center}
\includegraphics[scale=0.8]{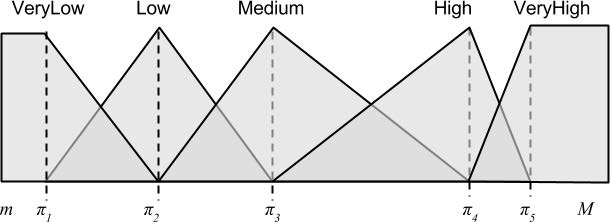}
\caption{Fuzzy sets construction using Fuzzy C-Means.}\label{partcmeans}
\end{center}
\end{figure}

\nd Another popular method is based on fuzzy clustering, \eg~using the so-called \emph{Fuzzy C-Means} algorithm~\cite{Bezdek81} (see figure~\ref{partcmeans}). 
Fuzzy C-Means extends K-Means  to accommodates graded membership: essentially, we create five clusters, $c_1, \ldots, c_5$, from the clusters $c_1, \ldots, c_5$ we take the centroids $\pi_1, \ldots, \pi_5$, and then
build the fuzzy sets from the centroids.

The standard fuzzy set operations are defined for any $x\in X$  as in \refeq{sand} and \refeq{sor}. Note also that the set inclusion defined as in \refeq{ssincl} is indeed crisp in the sense that either $A \subseteq B$ or $A \not\subseteq B$.

\paragraph{Norm-Based Fuzzy Set Operations.}
Standard fuzzy set operations are not the only ones that can be conceived to be suitable to generalise the classical Boolean operations. For each of the three types of operations there is a wide class of plausible fuzzy version. The most notable ones are characterised by the so-called class of \emph{t-norms} $\otimes$ (called \emph{triangular norms}), \emph{t-conorms} $\oplus$ (also called \emph{s-norm}), and \emph{negation} $\ominus$ (see, \eg~\cite{Klement00}). An additional operator is used to define set inclusion (called \emph{implication} $\impf$). Indeed, the \emph{degree of subsumption} between two fuzzy sets $A$ and $B$, denoted $A \sqsubseteq B$, is defined as $\inf_{x \in X} A(x) \impf B(x)$, where $\impf$ is an implication function.

An important aspect of such functions  is that they satisfy some properties that one expects to hold (see Tables~\ref{tabprop} and~\ref{axiom-imp-neg}). Usually, the implication function $\impf$ is defined as \emph{r-implication}, that is,  
\[
a \impf b = \sup\,\{c\mid a \otimes c \leq b\} \ .
\]

\begin{table}
\caption{Properties for t-norms and s-norms.} \label{tabprop}
\begin{center}{\small
\begin{tabular}{lll}\hline
Axiom Name & T-norm & S-norm \\ \hline
{Tautology / Contradiction} &
$a \otimes 0 = 0$ &
$a \oplus 1 = 1$
\\
{Identity} &
$a \otimes 1 = a$ &
$a \oplus 0 = a$
\\ 
{Commutativity} &
$a \otimes b = b \otimes a$ &
$a \oplus b = b \oplus a$
\\ 
{Associativity} & 
$(a \otimes b) \otimes c =
a \otimes (b \otimes c)$ & 
$(a \oplus b) \oplus c =
a \oplus (b \oplus c)$
\\ 
{Monotonicity} & 
if $b \le c$, then $a \otimes b \le
a \otimes c$ & 
if $b\le c$, then $a \oplus b \le
a \oplus c$ 
\\\hline
\end{tabular}}
\end{center}
\end{table}

\begin{table}
\caption{\label{axiom-imp-neg} Properties  for implication and negation functions.}
\begin{center}{\small
\begin{tabular}{llr}\hline
Axiom Name & Implication Function & Negation Function \\ \hline
{Tautology / Contradiction} &
$0 \impf b = 1$, \ $a \impf 1 = 1$, \ $1 \impf 0 = 0$ &
${\ominus}\,0 = 1$, \ ${\ominus}\, 1 = 0$
\\
{Antitonicity} & 
if $a \leq b$, then $a \impf c \geq b\impf c$ & 
\!\!\!\!\!\!\!\!\!\!if $a \leq b$, then ${\ominus}\, a \geq {\ominus}\, b$
\\ 
{Monotonicity} & 
if $b \leq c$, then $a \impf b \leq a \impf c$ & 
\\\hline
\end{tabular}}
\end{center}
\end{table}

\nd Of course, due to  commutativity, $\otimes$ and $\oplus$ are monotone also in the first argument. We say that $\otimes$ is \emph{indempotent} if $a \otimes a = a$,  for any $a\in [0,1]$. 
For any $a\in [0,1]$, we say that a negation function $\ominus$ is \emph{involutive} iff $\ominus \ominus a = a$ .  Salient negation functions are:
\begin{description}
\item[Standard  or \luks~negation:] $\ominus_{l} a = 1-a$;
\item[\godel~negation:] $\ominus_{g} a$ is $1$ if $a=0$, else is $0$.
\end{description}

\nd Of course,  \L ukasiewicz negation is involutive, while G\"{o}del negation is not.

Salient t-norm functions are:
\begin{description}
\item[\godel~t-norm:] $a \otimes_{g}b = \min(a,b)$;
\item[Bounded difference or \luks~t-norm:] $a \otimes_{l} b = \max(0, a+ b -1)$;
\item[Algebraic product or product t-norm:] $a \otimes_{p} b = a \cdot b$;
\item[Drastic product:] 
$a \otimes_{d} b  = \left \{  
\begin{array}{ll}
0 & \mbox{when } (a,b) \in [0,1[ \times [0,1[\\
\min(a,b) & \mbox{otherwise}
\end{array}
\right .$
\end{description}

\nd Salient s-norm functions are:
\begin{description}
\item[\godel~s-norm:] $a \oplus_{g}b = \max(a,b)$;
\item[Bounded sum or \luks~s-norm:] $a \oplus_{l} b = \min(1, a+ b)$;
\item[Algebraic sum or product s-norm:] $a \oplus_{p} b = a + b - ab$;
\item[Drastic sum:] 
$a \oplus_{d} b  = \left \{  
\begin{array}{ll}
1 & \mbox{when } (a,b) \in ]0,1] \times ]0,1]\\
\max(a,b) & \mbox{otherwise}
\end{array}
\right .$
\end{description}

\nd We recall that the following important properties can be shown about t-norms and s-norms.

\begin{enumerate}
\item There is the following ordering among  t-norms ($\otimes$ is any t-norm):
\begin{eqnarray*}
&& \otimes_{d} \leq \otimes \leq \otimes_{g} \\
&& \otimes_{d} \leq \otimes_{l} \leq \otimes_{p} \leq \otimes_{g} \ .
\end{eqnarray*}

\item The only idempotent t-norm is $\otimes_{g}$.
\item The only  t-norm satisfying $a \otimes a = 0$ for all $a \in [0,1[$ is $\otimes_{d}$.
\item There is the following ordering among  s-norms ($\oplus$ is any s-norm):
\begin{eqnarray*}
&& \oplus_{g} \leq \oplus \leq \oplus_{d} \\
&& \oplus_{g} \leq \oplus_{p} \leq \oplus_{l} \leq \oplus_{d} \ .
\end{eqnarray*}

\item The only idempotent s-norm is $\oplus_{g}$.
\item The only  s-norm satisfying $a \oplus a = 1$ for all $a \in ]0,1]$ is $\oplus_{d}$.
\end{enumerate}

\nd The  \emph{dual s-norm} of $\otimes$ is defined as 
\begin{equation} \label{eqtsnorm}
a \oplus b = 1 - (1-a) \otimes (1-b) \ . 
\end{equation}

\nd Some  {t-norms}, {s-norms},  {implication functions}, and {negation functions} are shown in Table~\ref{tabnorm}.
One usually distinguishes three different sets of fuzzy set operations (called fuzzy logics),  namely, \L{}ukasiewicz, G{\"o}del, and Product logic; the popular Standard Fuzzy Logic (SFL) is a sublogic of  \L{}ukasiewicz logic as $ \min(a,b) = a \otimes_{l} (a \impf_{l} b)$ and $\max(a,b) = 1 - \min(1-a, 1-b)$. The importance of these three logics is due to the Mostert--Shields theorem~\cite{Mostert57} that states that any continuous t-norm can be obtained as an ordinal sum of these three (see also~\cite{HajekP98}).

\begin{table}
\caption{Combination functions of various fuzzy logics.} \label{tabnorm}
\begin{center}{\small
\begin{tabular}{c@{\ \ \ \ \ \ \,}c@{\ \ \ \ \ \ \,}c@{\ \ \ \ \ \ \,}c@{\ \ \ \ \ \ \,}c} \hline
     & \mbox{\L{}ukasiewicz Logic} &  \mbox{G{\"{o}}del Logic}  & \mbox{Product Logic} & \mbox{SFL} \\ \hline
    $a \otimes b$ & $\max(a+b-1,0)$ & $\min(a,b)$ & $a\cdot b$  & $\min(a,b)$ \\
    $a \oplus b$ &  $\min(a+b,1)$ & $\max(a,b)$ & $a+b-a\cdot b$ & $\max(a,b)$ \\ 
    $a \impf b$ & $\min(1-a+b,1)$ &  $\begin{cases} 1 & \mbox{\!if } a \leq b\\
                                 b & \mbox{\!otherwise}\end{cases}$ & $\min(1,b/a)$ & $\max(1-a,b)$ \\[2.5ex] 
        ${\ominus}\, a$ & $1-a$ &  $\begin{cases} 1 & \mbox{\!if } a =0\\
                                 0 & \mbox{\!otherwise}\end{cases}$
	 & $\begin{cases} 1 & \mbox{\!if } a =0\\ 
                                 0 & \mbox{\!otherwise}\end{cases}$
     & $1-a$ \\ \hline

\end{tabular}}
\end{center}
\end{table}

\begin{table}
\caption{Some additional properties of combination functions of various fuzzy logics.} \label{tabpropnorm}
\[{\small
\begin{array}{ccccc} \hline
\text{Property} & \mbox{\,\L{}ukasiewicz Logic\,} &  \mbox{\,G{\"{o}}del Logic\,}  & \mbox{\,Product Logic\,} & \mbox{\,SFL\,} \\ \hline 
x \otimes {\ominus}\, x  =  0  & + & - & - &-\\ 
x\oplus {\ominus}\, x =  1  & +  & - & - &- \\  
x\otimes x=  x&- & + & - &+ \\  
x\oplus x=  x&- & + & - &+ \\  
{\ominus}\, {\ominus}\, x=  x&+ & - & - &+ \\  
x \impf y = {\ominus}\, x \oplus y&+ & - & - &+ \\  
{\ominus}\, (x \impf y) = x \otimes {\ominus}\, y &+ & - & - &+ \\  
{\ominus}\, (x\otimes y) = {\ominus}\, x \oplus {\ominus}\, y&+  & + & + &+ \\   
{\ominus}\, (x\oplus y) = {\ominus}\, x \otimes {\ominus}\, y&+  & + & + &+ \\  \hline
\end{array}
}\]
\end{table}

The implication $x \impf y = \max(1-x,y)$ is called {\em Kleene-Dienes implication} in the 
fuzzy logic literature. Note that we have the following inferences: let $a \geq n$ and $a \impf b \geq m$. 
Then, under Kleene-Dienes implication, we infer that if $n > 1-m$ then $b \geq m$. Under r-implication 
relative to a t-norm $\otimes$, we infer that $b \geq n\otimes m$. 

The \emph{composition} of  two fuzzy relations $R_{1}\colon X\times X \to [0,1]$ and $R_{2}\colon X\times X \to [0,1]$ is defined as $(R_{1} \circ R_{2})(x,z) = \sup_{y\in X} R_{1}(x,y) \otimes R_{2}(y,z)$. A~fuzzy relation $R$ is \emph{transitive} iff  $R(x,z) \ges (R\circ R)(x,z)$.

\paragraph{Fuzzy Modifiers.}
\emph{Fuzzy modifiers} are an interesting feature of fuzzy set theory. Essentially, a fuzzy modifier, such as $\mathtt{very}$, $\mathtt{more\_or\_less}$, and $\mathtt{slightly}$, apply to fuzzy sets to change their membership function. 

Formally, a \emph{fuzzy modifier} $m$ represents a function 
\[
f_{m}\colon\unit \to \unit \ .
\]

\nd  For example, we  may define $f_\mathtt{very}(x) \eqs  x^{2}$ and $f_\mathtt{slightly}(x) =  \sqrt x$. In this way, we may express the fuzzy set of very heavy rain by applying the modifier $very$ to the fuzzy membership function of ``heavy rain'' \ie
\[
\chi_{\mathtt{very \, heavy rain}}(x) = f_{\emph{very}}(\chi_{\mathtt{ heavy rain}}(x)) = (\chi_{\mathtt{ heavy rain}}(x))^{2} =  (rs(5,7.5)(x))^{2} \ .
\]

\begin{figure}[t]
\begin{center}
\includegraphics[scale=0.7]{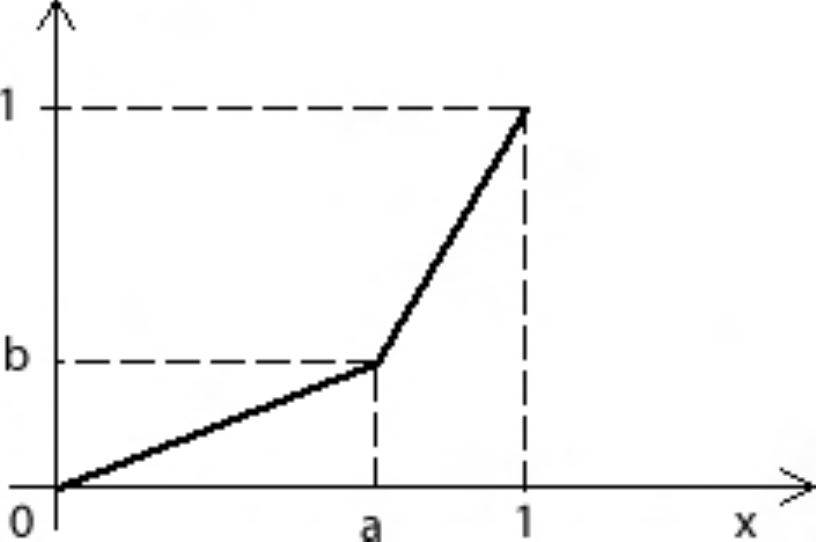}
\caption{Linear modifier $lm(a,b)$.}  \label{fig:lm}
\end{center}
\end{figure}

\nd A typical shape of modifiers is the so-called \emph{linear modifiers}, as illustrated in \reffig{fig:lm}.
Note that such a modifier can be parameterized by means of one parameter $c$ only, \ie~$lm(a,b) = lm(c)$, where
$a = c/(c+1) \ , \  b=1/(c+1)$.

\subsection{Mathematical Fuzzy Logic Basics}
\label{sect:mfuzzy}
\nd  We recap here briefly that in \emph{Mathematical Fuzzy Logic}~\cite{HajekP98}, the convention prescribing that a statement is either true or false is
changed and is a matter of degree measured on an ordered scale that
is no longer $\{0, 1\}$, but  $[0,1]$. This degree is called {\em
degree of truth} of the logical statement $\phi$ in the interpretation $\I$.
\emph{Fuzzy statements} have the form $\fuzzyg{\phi}{r}$, where $r \ins [0,1]$ (see,
\eg~\cite{Haehnle01,HajekP98}) and $\phi$ is a statement, which
encodes that the degree of truth  of $\phi$ is {\em greater or equal $r$}.
A \emph{fuzzy interpretation} $\I$ maps each basic statement $p_i$
into $[0,1]$ and is then extended inductively to all statements:
\begin{equation} 
\begin{array}{lcl}
\I(\phi \andf \psi) & = & \I(\phi) \otimes \I(\psi) \\
\I(\phi \orf \psi) & = & \I(\phi)\oplus \I(\psi) \\
\I(\phi \limpf \psi) & = & \I(\phi) \impf \I(\psi) \\
\I(\phi \liff \psi) & = &  \I(\phi \limpf \psi) \otimes \I(\psi \limpf \phi) \\
\I(\notf \phi) & = & {\ominus}\,\I(\phi) \\
\I(\exists x. \phi) &  = &  \sup_{a \in \highi{\Delta}}\I_{x}^{a}(\phi) \\
\I(\forall x. \phi) &  =  & \inf_{a \in \highi{\Delta}}\I_{x}^{a}(\phi) \ ,
\end{array}
\end{equation}
\nd where $\highi{\Delta}$ is the domain of
$\I$, and $\otimes$, $\oplus$, $\impf$, and $\ominus$ are the  \emph{t-norms}, \emph{t-conorms}, \emph{implication functions}, a
\emph{negation functions} we have seen in the previous section.\footnote{The function $\I_{x}^{a}$ is as $\I$ except that $x$ is interpreted as $a$.} 

One may also consider  the following abbreviations:
\begin{eqnarray}
\phi \land_{g} \psi & \mydef & \phi \land (\phi \limpf \psi) \label{abbrf1}\\
\phi \lor_{g} \psi & \mydef & (\phi \limpf \psi) \limpf \phi) \land_{g} (\psi \limpf \phi) \limpf \psi) \label{abbrf2}\\
\notf_{\otimes}\phi & \mydef &  \phi \limpf 0 \ . \\
\end{eqnarray}

\nd In case $\impf$ is the r-implication based on $\otimes$, then $\land_{g}$ (resp. $\lor_{g}$) is interpreted as \godel~t-norm (resp. s-norm), while $\notf_{\otimes}$ is interpreted as the negation function related to $\otimes$.


A fuzzy interpretation $\I$  \emph{satisfies} a fuzzy statement $\fuzzyg{\phi}{r}$, or  $\I$ is a \emph{model}
 of $\fuzzyg{\phi}{r}$, denoted $\I\,{\models}\,\fuzzyg{\phi}{r}$, iff $\I(\phi) \geq r$. 
%
%
We say that $\I$ is a \emph{model} of $\phi$ if $\I(\phi) = 1$.
A \emph{fuzzy knowledge base} (or simply knowledge base, if clear from context) is a set of fuzzy statements and an interpretation $\I$  \emph{satisfies} (is a \emph{model} of) a knowledge base, denoted $\I \models \KB$, iff it satisfies each element in it. 

We say~$\fuzzyg{\phi}{n}$ is a {\em tight logical consequence} of a set of fuzzy statements $\KB$  iff $n$ is the infimum of~$\I(\phi)$ subject to all models~$\I$ of~$\KB$. Notice
that the latter is equivalent to $n\eqs \sup \,\{ r \mids \KB \,{\models}\,\fuzzyg{\phi}{r}\}$. $n$ is called the \emph{best entailment degree} of $\phi$ \wrt~$\KB$ (denoted $\bed{\KB}{\phi}$),  \ie
\begin{equation}
\bed{\KB}{\phi} = \sup \,\{ r \mids \KB \,{\models}\,\fuzzyg{\phi}{r}\} \ . \label{bedfl}
\end{equation}

\nd On the other hand, the \emph{best satisfiability degree}  of $\phi$ \wrt~$\KB$ (denoted $\bsd{\KB}{\phi}$)  is
\begin{equation}
\bsd{\KB}{\phi} =  \sup_{\I} \,\{ \I(\phi) \mids \I \models \KB \} \ . \label{bsdfl}
\end{equation}

\nd Of course, the properties of Table~\ref{tabpropnorm} immediately translate into equivalence among formulae. For instance, the following equivalences hold (in brackets we indicate the logic for which the equivalences holds)
\begin{eqnarray*}
\notf \notf \phi & \equiv & \phi  \ \ (\L) \\
\phi \land  \phi & \equiv & \phi \  \ (G) \\
\notf ( \phi \land \notf \phi) & \equiv & 1 \ \  (\L, G, \Pi) \\
 \phi \lor \notf \phi & \equiv & 1 \ \ (\L) \ .
\end{eqnarray*}


\begin{remark} \label{fsomeall}
Unlike the classical case, in general, we do not have that  $\forall x. \phi$ and $\notf \exists x. \notf \phi$ are equivalent. They are equivalent for \luks~logic and SFL, but are neither equivalent for \godel~nor for Product logic. For instance, under  \godel~negation, just consider an interpretation $\I$ with domain $\{a\}$ and 
$\I(p(a)) = u$, with $0 < u < 1$. Then $\I(\forall x. p(x)) = u$, while $\I(\notf \exists x. \notf p(x)) = 1$ and, thus, $\forall x. p(x) \not \equiv \notf \exists x. \notf p(x)$.


\end{remark}

\nd We refer the reader to~\cite{Straccia13} for an overview of reasoning algorithms for fuzzy propositional and First-Order Logics. 

\subsection{Conjunctive Queries} \label{conjq}

\paragraph{The classical case.} In case a KB is a classical knowledge base, a \emph{conjunctive query} is a rule-like expression of the form 
\begin{equation} \label{eqdlcq}
q(\vec{x}) \leftarrow \exists \vec{y}.\varphi(\vec{x},\vec{y})
\end{equation}

\nd where  the rule body $\varphi(\vec{x},\vec{y})$  is a conjunction\footnote{We use the symbol $``,''$ to denote conjunction in the rule body.} of  predicates  $P_{i}(\vec{z}_{i})$ ($1 \leq i \leq n$) and $\vec{z}_{i}$ is a vector of distinguished or non-distinguished variables. 

For instance, 
\[
q(x,y) \leftarrow AdultPerson(x), Age(x,y)
\]

\nd is a conjunctive query, whose intended meaning is to retrieve all adult people and their age.

Given a vector $\vec{x} = \tuple{x_1, \dots, x_k}$ of variables, a \emph{substitution} over
  $\vec{x}$ is a vector of individuals $\vec{t}$ replacing variables in $\vec{x}$ with individuals. Then,
given a query $q(\vec{x}) \leftarrow \exists \vec{y}.\varphi(\vec{x},\vec{y})$, and two substitutions $\vec{t},
  \vec{t'}$ over $\vec{x}$ and $\vec{y}$, respectively, the \emph{query instantiation} $\varphi(\vec{t}, \vec{t}')$ is derived from $\varphi(\vec{x},
  \vec{y})$ by replacing  $\vec{x}$ and $\vec{y}$ with $\vec{t}$ and   $\vec{t}'$, respectively.

We adopt here the following notion of entailment. Given a knowledge base $\KB$, a query $q(\vec{x}) \leftarrow \exists \vec{y}.\varphi(\vec{x},\vec{y})$, and a vector $\vec{t}$ of   individuals occurring in $\KB$, we say that $q(\vec{t})$ is \emph{entailed} by $\KB$, denoted $\KB \models q(\vec{t})$, \iff 
there is a vector $\vec{t}'$ of individuals occurring in $\KB$ such that in any two-valued model $\I$ of $\KB$,   $\I$ is a model of any atom in the query  instantiation $\varphi(\vec{t}, \vec{t}')$.

If $\KB \models q(\vec{t})$ then $\vec{t}$ is called a \emph{answer} to $q$. We call these kinds of answers also \emph{certain answers}.
The \emph{answer set} of $q$ \wrt~$\KB$ is  defined as 
\[
ans(\KB,q) = \{\vec{t} \mid \KB \models q(\vec{t}) \} \ .
\]

\paragraph{The fuzzy case.} Consider a new alphabet of \emph{fuzzy variables} (denoted $\Lambda$). To start with, a \emph{fuzzy query} is of the form
\begin{equation} \label{fuzzyqDL}
\fuzzyg{q(\vec{x})}{\Lambda} \leftarrow \exists \vec{y}\exists\mathbf{\Lambda}'.\varphi(\vec{x}, \Lambda,\vec{y},\vec{\Lambda}')
\end{equation}

\nd  in which $\varphi(\vec{x}, \Lambda,\vec{y},\vec{\Lambda}')$ is a conjunction (as for the crisp case, we use ``,'' as conjunction symbol) of fuzzy predicates and built-in predicates, $\vec{x}$ and $\Lambda$ are the distinguished variables,  $\vec{y}$ and $\vec{\Lambda}'$ are the vectors of \emph{non-distinguished variables} (existential quantified variables), and $\vec{x}$, $\Lambda$, $\vec{y}$ and $\vec{\Lambda}'$ are pairwise disjoint. Variable  $\Lambda$ and variables in $\vec{\Lambda}'$ can only appear in place of degrees of truth or built-in predicates. The query head contains at least one variable. 

For instance, the query
\[
\fuzzyg{q(x)}{s} \leftarrow \fuzzyg{SportsCar(x)}{s_{1}}, hasPrice(x,y), s\assign s_{1} \cdot ls(10000,15000)(y)
\]

\nd  has intended meaning to retrieve all cheap sports cars. Any answer $x$ is scored according to the product of being cheap and a sports car, were cheap is encode as the fuzzy membership function $ ls(10000,15000)$.

From a semantics point of view, given a fuzzy KB $\KB$, a query $\fuzzyg{q(\vec{x})}{ \Lambda} \leftarrow \exists \vec{y}\exists\mathbf{\Lambda}'.\varphi(\vec{x}, \Lambda, \vec{y},\vec{\Lambda}')$, a vector $\vec{t}$ of individuals occurring in $\KB$ and a  truth degree  $\lambda$ in $[0,1]$, we say that $\fuzzyg{q(\vec{t})}{\lambda}$ is
\emph{entailed}  by $\KB$, denoted $\KB \models \fuzzyg{q(\vec{t})}{\lambda}$, \iff there is a vector
$\vec{t}'$ of individuals occurring $\KB$ and a vector $\vec{\lambda}'$ of truth degrees in $[0,1]$  such that  for any model $\I$ of $\KB$, $\I$ is a model of all fuzzy atoms occurring in $\varphi(\vec{t}, \lambda, \vec{t}', \vec{\lambda}')$. If $\KB \models \fuzzyg{q(\vec{t})}{\lambda}$ then $\tuple{\vec{t}, \lambda}$ is called an \emph{answer} to $q$. The \emph{answer set}  of $q$ \wrt~$\KB$ is 
\[
\begin{array}{ll}
ans(\KB, q) = \{ \tuple{\vec{t}, \lambda} \mid \KB \models  \fuzzyg{q(\vec{t})}{\lambda}, \lambda \neq 0 
\mbox { and  } \\
\hspace*{0.5cm} \mbox { for any }  \lambda' \neq \lambda 
   \mbox { such that } \KB \models  \fuzzyg{q(\vec{t})}{\lambda'}, \lambda' \leq \lambda  \mbox { holds} \} \ .
\end{array}
\]

\nd That is, for any tuple $\vec{t}$, the truth degree $\lambda$ is as large as possible.  

\paragraph{Fuzzy queries with aggregation operators.}
We may extend conjunctive queries to disjunctive queries and to queries including aggregation operators as well.
Formally, let $\aggr$ be an aggregate function with 
\[
\aggr \in \{\mathsf{SUM}, \mathsf{AVG}, \mathsf{MAX},\mathsf{MIN}, \mathsf{COUNT},\oplus, \otimes\} 
\]
\nd then a query with aggregates is of the form\index{conjunctive query with aggregates!fuzzy DLs}
\begin{equation}\label{dqaggrDL}
\begin{array}{lcl}
  \fuzzyg{q(\vec{x})}{\Lambda} & \leftarrow & \exists \vec{y}\exists\mathbf{\Lambda}'.\varphi(\vec{x}, \vec{y},\mathbf{\Lambda}'),\\
                                   &            & \mathsf{GroupedBy(\vec{w})},\\
                                   &            &  \Lambda \assign\aggr[f(\vec{z})] \ ,
\end{array}
\end{equation}

\nd where $\vec{w}$ are variables in $\vec{x}$ or $\vec{y}$ and each variable in $\vec{x}$ occurs in $\vec{w}$ and any variable in $\vec{z}$ occurs in $\vec{y}$ or $\vec{\Lambda'}$.

From a semantics point of view, we say that $\I$  \emph{is a model of} (\emph{satisfies}) \index{model!fuzzy DL query} $\fuzzyg{q(\vec{t})}{ \lambda}$, denoted 
$\I \models \fuzzyg{q(\vec{t})}{ \lambda}$ \iff
\[
\begin{array}{l}
\lambda =   \aggr [\lambda_{1}, \ldots, \lambda_{k}]  \mbox{ where } g = \{ \tuple{\vec{t}, \vec{t}'_{1},\vec{\lambda}'_{1}}, \ldots , \tuple{\vec{t}, \vec{t}'_{k},\vec{\lambda}_{k}'} \}, \\
\hspace{1.3cm}\mbox{is a group of $k$ tuples with identical projection}\\
\hspace{1.3cm}\mbox{on the variables in } \vec{w}, \varphi(\vec{t}, \vec{t}'_{r},\vec{\lambda}'_{r}) \mbox{ is true in } \I \\
\hspace{1.3cm}\mbox {and } \lambda_{r} =f(\vec{\vec{t}}) \mbox{ where } \vec{\vec{t}} \mbox{ is the projection of }  \tuple{\vec{t}'_{r}, \vec{\lambda}'_{r}}\\
\hspace{1.3cm}\mbox{on the variables } \vec{z} \ . \\
\end{array}
\]

\nd Now, the notion of  $\KB \models \fuzzyg{q(\vec{t})}{ \lambda}$ is as usual: any model of $\KB$ is a model of $\fuzzyg{q(\vec{t})}{ \lambda}$.


The notion of answer and answer set of a disjunctive query is a straightforward extension of the ones for conjunctive queries.

\paragraph{Top-k Retrieval.}
As now each answer to a query has a degree of truth (\ie~\emph{score}), a basic inference problem that is of interest is the top-$k$ retrieval problem, formulated as follows.  

Given a fuzzy KB $\KB$, and a  query $q$, retrieve $k$ answers $\fuzzyg{\vec{t}}{\lambda}$ with maximal degree and rank them in decreasing order relative to the degree $\lambda$, denoted\index{top-k retrieval!fuzzy DLs}
\[
ans_{k}(\KB,q) = \mathtt{Top}_{k} \ ans(\KB,q) \ .
\]

\subsection{Annotation Domains} \label{anot}

We have seen that fuzzy statements extend statements with an \emph{annotation} $r\in[0,1]$. Interestingly, we may further generalise this by allowing a statement being annotated with a value 
$\lambda$ taken from a so-called \emph{annotation domain}~\cite{Buneman10,Lopes10a,Lopes10,Straccia10a,Zimmermann12},\footnote{The readers familiar with the annotated logic programming framework~\cite{Kifer92}, will notice the similarity of the approaches.} which allow to deal with several domains (such as, fuzzy, temporal, provenance) and their combination, in a uniform way. 
Formally, let us consider a non-empty set $L$. Elements in $L$ are our annotation values. For example, in a fuzzy setting, $L = [0,1]$, while in a typical temporal setting, $L$ may be time points or time intervals. In the annotation framework, an interpretation will map statements to elements of the annotation domain.  Now, an \emph{annotation domain}  is an idempotent, commutative semi-ring 
\[
D = \tuple{L,  \oplus, \otimes, \bot, \top} \ ,
\]
\nd where $\oplus$ is $\top$-annihilating~\cite{Buneman10}. That is, for $\lambda, \lambda_{i} \in L$

\begin{enumerate}
\item $\oplus$ is idempotent, commutative, associative;
\item $\otimes$ is commutative and associative;
\item $\bot \oplus \lambda = \lambda$, $\top \otimes \lambda = \lambda$, $\bot \otimes \lambda = \bot$, and $\top \oplus \lambda = \top$; 
\item $\otimes$ is  distributive over $\oplus$, \ie 
$\lambda_{1} \otimes (\lambda_{2} \oplus \lambda_{3}) =  (\lambda_{1} \otimes \lambda_{2}) \oplus (\lambda_{1} \otimes \lambda_{3})$; 
\end{enumerate}

\nd It is well-known that there is a natural partial order on any idempotent semi-ring: an annotation domain 
$D = \tuple{L,  \oplus, \otimes, \bot, \top}$ induces a partial order $\preceq$ over $L$ defined as:
\[
\lambda_{1} \preceq \lambda_{2} \mbox{ \ \iff  \ }  \lambda_{1} \oplus \lambda_{2} = \lambda_{2} \ .
\]
\nd The order $\preceq$ is used to express redundant/entailed/subsumed information. For instance, for temporal intervals, an
annotated statement $\fuzzyg{\phi}{[2000,2006]}$ entails $\fuzzyg{\phi}{[2003,2004]}$, as $[2003,2004]
\subseteq [2000,2006]$ (here, $\subseteq$ plays the role of $\preceq$).
  
\begin{remark}\label{rem4}
 $\oplus$ is used to combine information about the same statement.  For instance, in temporal logic, from
  $\fuzzyg{\phi}{[2000,2006]}$ and $\fuzzyg{\phi}{[2003,2008]}$, we infer $\fuzzyg{\phi}{[2000,2008]}$, as $[2000,2008] =
  [2000,$ $2006]\cup [2003,2008]$; here, $\cup$ plays the role of $\oplus$. In the fuzzy context, from $\fuzzyg{\phi}{0.7}$ and
  $\fuzzyg{\phi}{0.6}$, we infer $\fuzzyg{\phi}{0.7}$, as $0.7 = \max(0.7, 0.6)$ (here, $\max$ plays the role of
  $\oplus$).
\end{remark}

\begin{remark}\label{rem5}
$\otimes$ is used to model the ``conjunction'' of information. In fact, a $\otimes$ is a generalisation of
  boolean conjunction to the many-valued case. In fact, $\otimes$ satisfies also that
 \begin{enumerate}
 \item $\otimes$ is bounded: \ie $\lambda_{1} \otimes \lambda_{2} \preceq \lambda_{1}$.   

 \item $\otimes$ is $\preceq$-monotone, \ie~for $\lambda_{1} \preceq \lambda_{2}$, $\lambda \otimes \lambda_{1} \preceq \lambda \otimes \lambda_{2}$
 
\end{enumerate}
\nd   For instance, on interval-valued temporal logic, from $\fuzzyg{\phi}{[2000,2006]}$ and $\fuzzyg{\phi \limpf \psi}{[2003,2008]}$, we may infer
  $\fuzzyg{\psi}{[2003,2006]}$, as $[2003,2006] = [2000,2006] \cap [2003,2008]$; here, $\cap$ plays
  the role of $\otimes$. In the fuzzy context, one
  may chose any t-norm~\cite{HajekP98,Klement00}, \eg~product, and, thus, from $\fuzzyg{\phi}{0.7}$ and $\fuzzyg{\phi \limpf \psi}{0.6}$, we will infer $\fuzzyg{\psi}{0.42}$, as $0.42 = 0.7 \cdot 0.6)$ (here,
  $\cdot $ plays the role of $\otimes$). 
\end{remark}

\begin{remark}\label{remdistr}
Observe that the distributivity condition is used to guarantee that \eg~we obtain the same annotation $\lambda \otimes (\lambda_{2} \oplus \lambda_{3}) = (\lambda_{1} \otimes \lambda_{2}) \oplus (\lambda_{1} \otimes \lambda_{3})$ of  
$\psi$ that can be inferred from  
$\fuzzyg{\phi}{\lambda_{1}}$, 
  $\fuzzyg{\phi  \limpf \psi}{\lambda_{2}}$ and 
  $\fuzzyg{\phi \limpf \psi }{\lambda_{3}}$.
\end{remark}

\nd Note that, conceptually, in order to build an annotation domain, one has to:
\begin{enumerate}
\item determine the set of annotation values $L$ (typically a countable set\footnote{Note that one may use XML decimals in $[0,1]$ in place of real numbers for the fuzzy domain.}), 
identify the top and bottom elements;

\item define a suitable operations $\otimes$ and $\oplus$ that acts as ``conjunction'' and ``disjunction'' function, to support the intended inferences.

\end{enumerate}

\nd Eventually, \emph{annotated queries} are as fuzzy queries in which annotation variables and terms are used in place of fuzzy variables and values $r \in [0,1]$ instead. 
We refer the reader to~\cite{Zimmermann12} for more details about annotation domains.

\section{Fuzzy Logic and Semantic Web Languages} \label{fswl}

We have seen in the previous section how to ``fuzzyfy''  a classical language such as propositional logic and FOL, namely fuzzy statements are of the form $\fuzzyg{\phi}{r}$, where $\phi$ is a statement and $r \in[0,1]$.

 The natural extension to SWLs consists then in replacing $\phi$ with appropriate expressions belonging to the logical counterparts of SWLs, namely $\rhodf$, DLs and LPs, as we will illustrate next.

\subsection{Fuzzy RDFS} \label{frdf}

The basic ingredients of \emph{RDF} are \emph{triples} of the form 
$\triple{s,p,o}$,  such as  $\triple{umberto, likes, tomato}$, stating that \emph{subject} $s$ has \emph{property} $p$ with \emph{value} $o$.   In \emph{RDF Schema} (RDFS), which is an extension of RDF, additionally some special keywords may be used as properties to further improve the expressivity of the language. For instance we may also express that the class of 'tomatoes are a subclass of  the class of vegetables', $\triple{tomato, \subclass, vegetables}$, while Zurich is an instance of the class of cities, $\triple{zurich, \type, city}$.  

Form a computational point of view, one computes the so-called \emph{closure} (denoted $cl(\KB)$)  of a set of triples $\KB$. That is, one  infers all possible triples using inference rules~\cite{Marin04,Munoz07,RDFmt}, such as 
{\small
\[
 \frac{\triple{A, \subclass, B},  \triple{X, \type, A}}{\triple{X, \type, B}} 
\]}
{\small
\begin{quote}
``if $A$ subclass of $B$ and $X$ instance of $A$ then infer that $X$ is instance of $B$'',
\end{quote}
}
\nd and then store all inferred triples into a relational database to be used then for querying. We recall also that there also several ways to store the closure $cl(\KB)$ in a database (see~\cite{Abadi09,Ianni09}). Essentially, either we may store all the triples in table with three columns \emph{subject, predicate, object}, or we use a table for each predicate, where each table has two columns \emph{subject, object}. The latter approach seems to be better for query answering purposes.

In \emph{Fuzzy RDFS} (see~\cite{Straccia09f,Straccia13} and references therein),  triples are annotated with a degree of truth
in $[0,1]$. For instance, ``Rome is a big city to degree 0.8'' can be represented with $\fuzzyg{\triple{\term{Rome},
    \type, \term{BigCity}}}{0.8}$. More formally, \emph{fuzzy triples} are expressions of the form 
$\fuzzyg{\tau}{r}$, where $\tau$ is a RDFS triple (the truth value $r$ may be omitted and, in that case, the value $r=1$ is assumed).

The interesting point is that from a computational point of view the inference rules parallel those for ``crisp'' RDFS: indeed, the rules are of the form
\begin{equation} \label{eqnn}
  \frac{
      \fuzzyg{\tau_{1}}{r_{1}},\ \ldots,\ \fuzzyg{\tau_{k}}{r_{k}}, \{\tau_{1}, \ldots, \tau_{k}\} \vdash_{\mathsf{RDFS}} \tau
    }
    {
      \fuzzyg{\tau}{\bigotimes_{i} r_{i}}
    }
\end{equation}

\nd Essentially, this rule says that if a classical RDFS triple $\tau$ can be inferred by applying a classical RDFS
inference rule to triples $\tau_{1}, \ldots, \tau_{k}$ (denoted $\{\tau_{1}, \ldots, \tau_{k}\} \vdash_{\mathsf{RDFS}}
\tau$), then the truth degree of $\tau$ will be $\bigotimes_{i} r_{i}$.

As a consequence, the rule system is quite easy to implement for current inference systems. Specifically, as for the crisp case, one may compute the closure $cl(\KB)$ of a set of fuzzy triples $\KB$, store them in a relational database and thereafter query the database. 

Concerning conjunctive queries, they are essentially the same as in Section~\ref{conjq}, where predicates are replaced with triples.
For instance, the query
\begin{equation} \label{qfrde}
\fuzzyg{q(x)}{s} \leftarrow \fuzzyg{\triple{x, \type, \term{SportsCar}}}{s_{1}}, \triple{x,  \term{hasPrice},  y}, s= s_{1} \cdot cheap(y)
\end{equation}
\nd where \eg~$cheap(y) = ls(10000, 15000)(y)$, has intended meaning to retrieve all cheap sports car. Then, any answer is scored according to the product of being cheap and a sports car.

\subsubsection{Annotation domains \& RDFS.} \label{ardfs}
The generalisation to annotation domains is conceptual easy, as now one may replace truth degrees with annotation terms taken from an appropriate domain.
For further details see~\cite{Zimmermann12}.

\subsection{Fuzzy DLs}
\emph{Description Logics} (DLs)~\cite{Baader03a} are the logical counterpart of the family of OWL languages. So, to illustrate the basic concepts of fuzzy OWL, it suffices to show the fuzzy DL case (see~\cite{Bobillo15b,Lukasiewicz08a,Straccia13}, for a survey).
We recap that the basic ingredients are the descriptions of classes, properties, and their instances, such as 
\begin{itemize}
\item $\cass{a}{C}$, such as $\cass{\ind{a}}{\cp{Person} \andc  \all \role{hasChild}.\cp{Femal}}$,
meaning that individual $a$ is an instance of concept/class $C$ (here $C$ is seen as a unary predicate);

\item $\rass{a}{b}{R}$, such as $\rass{\ind{tom}}{\ind{mary}}{\role{hasChild}}$,
meaning that the pair of individuals  $\tuple{a,b}$ is an instance of the property/role $R$ (here $R$ is seen as a binary predicate);

\item $C \impc D$, such as $\cp{Person} \impc  \all \role{hasChild}.\cp{Person}$, meaning that the class $C$ is a subclass of class $D$;

\end{itemize}

\nd So far, several \emph{fuzzy} variants of DLs have been proposed: they can be classified according to 
 
 \begin{itemize}
\item the description logic resp.\ ontology language that they generalize~\cite{Bobillo08c,Bobillo08b,Bobillo09e,Bobillo09b,Bobillo09a,Bobillo17,Borgwardt15c,Borgwardt16,Dubois06,Lukasiewicz06a,Lukasiewicz08c,Lukasiewicz07d,Lukasiewicz07,Lukasiewicz07e,Lukasiewicz08b,Lukasiewicz09,SanchezD04,SanchezD05,SanchezD06,Stoilos07a,Straccia98,Straccia05d,Straccia05e,Straccia06c,Straccia08,Venetis07,Yen91a};
\item the allowed fuzzy constructs~\cite{Bobillo08a,Bobillo10b,Bobillo11b,Bobillo11d,Bobillo11c,Bobillo12,Bobillo13a,Bobillo13b,Hoelldobler06,Hoelldobler02,Hoelldobler05,Hoelldobler03,Hoelldobler04a,Hoelldobler04,Jiang11,Jiang10,Jiang10a,Jiang09,Jiang10b,Jiang09b,Jiang09a,Kang06,Mailis07,Straccia05d,Straccia09a,Tresp98};
\item the underlying fuzzy logic~\cite{Bobillo09,Bobillo07,Bobillo09c,HajekP05,HajekP06,Straccia04,Straccia06,Straccia06d}; 
\item their reasoning algorithms and computational complexity results~\cite{Baader11,Baader11a,Bobillo11a,Bobillo06,Bobillo08d,Bobillo08e,Bobillo09,Bobillo12a,Bobillo08,Bobillo11,Bobillo13,Bobillo14,Bobillo15a,Bobillo16a,Bobillo16c,Bobillo18,Bonatti03,Borgwardt12,Borgwardt12c,Borgwardt11,Borgwardt11b,Borgwardt11a,Borgwardt12d,Borgwardt12a,Borgwardt13,Borgwardt13a,Borgwardt13b,Borgwardt14,Borgwardt15,Borgwardt15a,Borgwardt15b,Borgwardt17,Borgwardt17a,Bou11,Cerami10,Cerami11a,Cerami11,Bobillo15,Pan07,Stoilos05a,Stoilos07b,Stoilos06,Straccia01,Straccia04d,Straccia06f,Straccia06i,Straccia07d,Straccia07c,Straccia07e,LiY06,Zhou12}.
 \end{itemize}


\nd In general, fuzzy DLs allow  expressions of the form $\fuzzyg{\cass{a}{C}}{r}$, stating that $a$ is an instance of concept/class $C$ with degree at least $r$, \ie~the FOL formula $C(a)$ is true to degree at least $r$. Similarly, $\fuzzyg{C_{1} \impc C_{2}}{r}$ states a vague subsumption relationships. Informally, $\fuzzyg{C_{1} \impc C_{2}}{r}$ dictates that the FOL formula $\forall x. C_{1}(x) \limpf C_{2}(x)$ is  true to degree at least $r$. Essentially, \emph{fuzzy DLs} are then obtained by interpreting the statements as fuzzy FOL formulae and attaching a weight $n$ to DL statements, thus, defining so \emph{fuzzy DL statements}.

\begin{example} \label{carsfDL}
Consider the following background knowledge about cars:
\[
\begin{array}{rcl}
Car  & \impc & \some HasPrice.Price \\
Sedan & \impc & Car \\
Van & \impc & Car \\
CheapPrice  &\impc&  Price \\
ModeratePrice & \impc  &Price\\ 
ExpensivePrice  &\impc & Price \\
\langle CheapPrice &  \impc & ModeratePrice, 0.7 \rangle \\
\langle ModeratePrice & \impc &  ExpensivePrice, 0.4\rangle \\
CheapCar & \defc &  Car \andc \some HasPrice.CheapPrice \\
ModerateCar  & \defc &  Car \andc \some HasPrice.ModeratePrice \\
ExpensiveCar & \defc & Car \andc \some HasPrice.ExpensivePrice \\
\end{array}
\]

\nd Essentially, the vague concepts here are $CheapPrice, ModeratePrice$, and $ExpensivePrice$ and the graded GCIs declare to which extent there is a relationship among them.

The facts about two specific cars $a$ and $b$ are encoded with:
\[
\begin{array}{l}
\fuzzyg{\cass{a}{Sedan \andc \some HasPrice.CheapPrice}}{0.7} \\
\fuzzyg{\cass{b}{Van \andc \some HasPrice.ModeratePrice}}{0.8} \ .
\end{array}
\]
\nd So, $a$ is a sedan having a cheap price, while $b$ is a van with a moderate price.

Under~\godel~semantics it can be shown that
\begin{eqnarray*}
\K & \models  & \fuzzyg{\cass{a}{ModerateCar}}{0.7} \\
\K & \models  & \fuzzyg{\cass{b}{ExpensiveCar}}{0.4}  \ .
\end{eqnarray*}


\end{example}

\nd From a decision procedure point of view, a popular approach consists of a set of inference rules that generate a set of in-equations (that depend on the t-norm and fuzzy concept constructors) that have to be solved by an operational research solver (see, \eg~\cite{Bobillo09c,Straccia05d}). An informal rule example is as follows:
\begin{quote}
``If individual $a$ is instance of the class intersection $C_{1} \andc C_{2}$ to degree greater or equal to $x_{\cass{a}{C_{1} \andc C_{2}}}$,\footnote{For a fuzzy DL formula $\phi$ we consider a variable $x_{\phi}$ with intended meaning: the degree of truth of $\phi$ is greater or equal to $x_{\phi}$.} then $a$ is instance of $C_{i}$ ($i=1,2$) to degree greater or equal to $x_{\cass{a}{C_{i}}}$, where additionally the following in-equation holds: 
\[
x_{\cass{a}{C_{1} \andc C_{2}}} \leq x_{\cass{a}{C_{1}}} \otimes x_{\cass{a}{C_{2}}} \mbox{  .'' }
\]
\end{quote}

\nd Concerning conjunctive queries, they are essentially the same as in Section~\ref{conjq}, where predicates are replaced with unay and binary predicates.
 For instance, the fuzzy DL analogue of the RDFS query~(\ref{qfrde}) is
\begin{equation} \label{qfDLe}
\fuzzyg{q(x)}{s} \leftarrow \fuzzyg{\term{SportsCar}(x)}{s_{1}}, \term{HasPrice}(x, y), s\assign s_{1} \cdot cheap(y) \ .
\end{equation}

\paragraph{Applications.}
Fuzzy set theory and fuzzy logic~\cite{Zadeh65} have proved to be
suitable formalisms to handle fuzzy knowledge. Not surprisingly, 
\emph{fuzzy ontologies} already emerge as useful in several applications,
such as information
retrieval~\cite{Andreasen09,Calegari08,LiuC12,Straccia07a,Straccia08b,Wallace09,ZhangL05},
recommendation
systems~\cite{Carlsson12,Lee10,Carlsson13,Yaguinuma13},
image interpretation~\cite{Dasiopoulou10,Dasiopoulou09,Dasiopoulou10a,Meghini01,Stoilos05,Straccia99a,Straccia00}, the
Semantic Web and the Internet~\cite{URRSW08,Quan06,Sanchez06}, ambient
intelligence~\cite{Rodriguez14a,Rodriguez13,LiuC10,Rodriguez14},
ontology merging~\cite{Chen11,Todorov14},
matchmaking~\cite{Agarwal05,Colucci10,Ragone08c,Ragone07a,Ragone07,Ragone08,Ragone09,Straccia09c,Straccia09d}, decision
making~\cite{Straccia09b}, summarization~\cite{Lee05},
robotics~\cite{Eich14,Eich13}, 
machine learning~\cite{Lisi13,Lisi13a,Lisi14a,Lisi15,Lisi11a,Lisi11,Lisi13b,Lisi14,Straccia15}
and many
others~\cite{Balaj13,Aquin06,Fernandez10,IglesiasI11,Konstantopoulos09,Letia09,LiuO04,Cruz12,Quan06a,Rodger12,Slavicek13,Straccia09a}.

\paragraph{Representing Fuzzy OWL Ontologies in OWL.} \label{fowlowl}
OWL~\cite{OWL} and its successor OWL 2~\cite{OWL2JWS,OWL2} are  standard W3C languages for defining and instantiating Web ontologies whose logical counterpart are classical DLs. So far, several fuzzy extensions of DLs exists and some fuzzy DL reasoners have been implemented, such as \textsc{fuzzyDL}~\cite{Bobillo08a,Bobillo16}, \textsc{DeLorean}~\cite{Bobillo08c}, \textsc{Fire}~\cite{Fire08,Stoilos06a}, \textsc{SoftFacts}~\cite{Straccia10b}, {\em GURDL}~\cite{Haarslev07}, {\em GERDS}~\cite{Habiballa07},  {\em YADLR}~\cite{Konstantopoulos07}, {\em FRESG}~\cite{Wang09} and \textsc{DLMedia}~\cite{Straccia10,Straccia08b}. 
Not surprisingly, each reasoner uses its own fuzzy DL language for representing fuzzy ontologies and, thus, there is a need for a standard way to represent such information.
A first possibility would be to adopt as a standard one of the fuzzy
extensions of the languages OWL and OWL 2 that have been proposed,
such as~\cite{Gao05a,Stoilos10a,Stoilos07a}. However, as it is not expected that a fuzzy OWL extension will become a W3C proposed standard in the near future,  \cite{Bobillo09d,Bobillo10,Bobillo11c}  identifies the syntactic differences that a fuzzy ontology language has to cope with, and proposes to use OWL 2 \emph{itself} to represent fuzzy ontologies~\cite{FuzzyOWL2}.

\subsubsection{Annotation domains \& OWL.} \label{aowl}
The generalisation to annotation domains is conceptual easy, as now one may replace truth degrees with annotation terms taken from an appropriate domain (see, \eg~\cite{Borgwardt11,Borgwardt11a,Straccia06d}.

\subsection{Fuzzy Rule Languages} \label{frif}

The foundation of the core part of rule languages is \emph{Datalog}~\cite{Ullman89}, \ie~a Logic Programming Language (LP)~\cite{Lloyd87}.
In  LP, the management of imperfect information has attracted the  attention of many researchers and numerous frameworks have been proposed. Addressing all of them is almost impossible, due to both the large number of works published in this field (early works date back to early 80-ties~\cite{Shapiro83}) and the different approaches proposed (see, \eg~\cite{Straccia08a}). 

\nd Below a list of references.\footnote{The list of references is by no means intended to be all-inclusive. The author apologises both to the authors and with the readers for all the relevant works, which are not cited here.}

\begin{description}
%

\item[Fuzzy set theory:]~\cite{Baldwin95,Baldwin98a,Bueno97,Cao00,Chortaras06,Chortaras07,Chortaras07a,Ebrahim01,Guller02,Guller01,Hinde86,Ishizuka85,Klawonn94,Smets89,Martin87,Mukaidono96,Mukaidono89,Paulik96,Rhodes95,Sessa02,Shapiro83,ShenZ88,SubramanianVS87,vanEmden86,Vojtas01,Vojtas96,Vojtas04,Wagner98,Yasui95}

\item[Multi-valued  logic:]~\cite{Calmet96,Damasio04c,Damasio04,Damasio04a,Damasio06b,Damasio08,Damasio98,Damasio01,Damasio01a,Damasio04b,Damasio06a,Damasio06c,Damasio07a,Denecker00a,Denecker01b,Denecker00,Denecker01,Denecker02,Fitting93,Fitting02,Fitting85,Fitting87,Fitting91,Haehnle91,Khamsi96,Khamsi97,Kifer88,Kifer92,Kullmann04,Lakshmanan94,Lakshmanan97,Lakshmanan01,Loyer02d,Loyer02b,Loyer03c,Loyer03d,Loyer04a,Loyer05,Loyer06,Lu96,Lu97,Lukasiewicz07d,Lukasiewicz07a,Lukasiewicz07e,Lukasiewicz08b,Lukasiewicz10,Madrid13,Majkic04,Majkic05a,Majkic05,Marek00,Mateis99,Mateis00,Medina04,Medina01,Medina01a,Medina04a,Ragone08c,Ragone07a,Ragone07,Ragone08,Ragone09,Rounds01,Schroeder01,Schroeder02,Schroeder04,Straccia05b,Straccia05c,Straccia06b,Straccia06c,Straccia06e,Straccia06h,Straccia06,Straccia07b,Straccia07,Straccia08,Straccia14,Straccia12a,Straccia09,TurnerH94}

\end{description}

\nd Basically~\cite{Lloyd87}, a Datalog program $\lp$ is made out by a set of rules and a set of facts. \emph{Facts} are ground \emph{atoms} of the form $P(\vec{c})$. On the other hand rules are similar as  conjunctive  queries and are of the form 
\[
A(\vec{x}) \leftarrow \exists \vec{y}.\varphi(\vec{x},\vec{y}) \ ,
\]
\nd where  $\varphi(\vec{x},\vec{y})$ is a conjunction of $n$-ary predicates. 
%
A \emph{query} is a rule and the \emph{answer set} of a query $q$ \wrt~a set $\KB$ of facts and rules is the set of tuples $\vec{t}$ such that there exists $\vec{t}'$ such that the instantiation $\varphi(\vec{t},\vec{t}')$ of the query body is true in \emph{minimal model} of $\KB$, which is guaranteed to exists. 

In the \emph{fuzzy} case, rules and facts are as for the crisp case, except that now a predicate is annotated.
An example of fuzzy rule defining good hotels may be the following:
{\small
\begin{eqnarray} \label{fgh}
\fuzzyg{GoodHotel(x)}{s} &  \leftarrow &
Hotel(x), 
\fuzzyg{Cheap(x)}{s_{1}}, 
\fuzzyg{CloseToVenue(x)}{s_{2}}, \nonumber \\ 
&& \fuzzyg{Comfortable(x)}{s_{3}},  s\assign~0.3\cdot s_{1} + 0.5 \cdot s_{2} + 0.2 \cdot s_{3}
\end{eqnarray}
}

\nd A \emph{fuzzy query} is a fuzzy rule and, informally, the \emph{fuzzy answer set} is the ordered set of weighted tuples $\tuple{\vec{t},s}$ such that all the fuzzy atoms in the rule body are true in the minimal model and $s$ is the result of the scoring function $f$ applied to its arguments. The existence of a minimal is guaranteed if the scoring functions in the query and in the rule bodies are \emph{monotone}~\cite{Straccia08a}.

We conclude by saying that most works deal with logic programs without negation and some may provide some technique to answer queries in a top-down manner, as \eg~\cite{Damasio04,Kifer92,Lakshmanan01,Straccia05c,Vojtas01}. Deciding whether a wighted tuple $\tuple{\vec{t},s}$ is the answer set is undecidable in general, though is decidable if the truth space is finite and fixed a priory, as then the minimal model is finite.

Another rising problem is the problem to compute the top-k ranked answers to a query, without computing the score of all answers. This allows to answer queries such as ``find the top-k closest hotels to the conference location''. Solutions to this problem can be found in~\cite{Lukasiewicz07e,Straccia06h,Straccia07}.
%
%
\subsubsection{Annotation domains \& Rule Languages.} \label{arif}
%
The generalisation of fuzzy rule languages to the case in which an annotation $r \in [0,1]$ is replaced with an annotation value $\lambda$ taken from an annotation domain is straightforward and proceeds as for the other SWLs. 

%
%

%
%



\end{document}